\documentclass[11pt]{article}

\usepackage[preprint]{neurips_2024}

\usepackage{times}
\usepackage{latexsym}
\usepackage[T1]{fontenc}
\usepackage[utf8]{inputenc}
\usepackage{microtype}
\usepackage{inconsolata}
\usepackage{graphicx}
\usepackage{amsmath,amsfonts,amssymb,amsthm,mathtools}
\usepackage{bbm}
\usepackage{booktabs}
\usepackage{multirow}
\usepackage{subcaption}
\usepackage{xcolor}
\usepackage{algorithm}
\usepackage{algpseudocode}
\usepackage{pifont}
\usepackage{url}
\usepackage{enumitem}
\usepackage[nolist,nohyperlinks]{acronym}

\usepackage{natbib}

\newcommand{\HPO}{\textsc{HPO}}
\newcommand{\AHPO}{\textsc{A-HPO}}
\newcommand{\VHPO}{\textsc{V-HPO}}
\newcommand{\GRPO}{\textsc{GRPO}}
\newcommand{\GSPO}{\textsc{GSPO}}
\newcommand{\SAPO}{\textsc{SAPO}}

\newcommand{\E}{\mathbb{E}}

\newcommand{\clip}{\mathrm{clip}}

\def\cal#1{\mathcal{#1}}

\makeatletter
\newcommand{\substackleft}[1]{%
  \vcenter{%
    \Let@ \restore@math@cr \default@tag
    \baselineskip\fontdimen10 \scriptfont\tw@
    \advance\baselineskip\fontdimen12 \scriptfont\tw@
    \lineskip\thr@@\fontdimen8 \scriptfont\thr@@
    \lineskiplimit\lineskip
    \ialign{$\m@th\scriptstyle##$&$\m@th\scriptstyle{}##$\hfil\crcr
      #1\crcr
    }%
  }%
}
\makeatother

\title{HPO: Hysteretic Policy Optimization for Stable and Efficient Training under Sparse-Reward Regime}

\author{
Mohamed Sana$^{\dagger}$, Nicola Piovesan$^{\dagger}$, Antonio De Domenico$^{\dagger}$, Fadhel Ayed$^{\dagger}$, Haozhe Zhang$^*$\\
$^{\dagger}$Paris Research Center, Huawei Technologies, Boulogne-Billancourt, France\\ 
$^*${Huawei Technologies, China}\\
}

\begin{document}

\maketitle

\begin{abstract}
We investigate a narrow but common failure mode of GRPO-style reinforcement learning in the context of sparse verifiable rewards: early updates contain more responses with negative advantages than those with positive advantages, while response-level length normalization ties the magnitude of the update to the length of the output. We propose \textbf{Hysteretic Policy Optimization (HPO)}, a minimal modification of GRPO that reduces the weight of negative-advantage updates and replaces per-response length normalization with mean-length normalization. We further introduce \textbf{Adaptive HPO (A-HPO)}, which sets the hysteretic weight based on batch-level advantage-sign statistics, thereby removing the need for tuning a fixed hysteretic weight. In our TeleLogs and Countdown experiments, A-HPO improves the reward per update compared to GRPO, with the largest gains in early sparse reward regimes. On TeleLogs, A-HPO achieves a final reward of 0.84, outperforming SAPO by $5\%$, GSPO by $11\%$, and GRPO by $15\%$, while maintaining a comparable response-length. On Countdown, A-HPO achieves the largest gains in initial and most difficult configurations across 1.5B--7B models. Ablation studies on the hysteretic weight show that the gains of A-HPO come from better balancing the contributions of positive and negative advantages compared to positive-only or fully symmetric updates.

\end{abstract}

\section{Introduction}\label{sec:intro}

\begin{figure}[htb]
  \centering
  \includegraphics[width=0.75\textwidth]{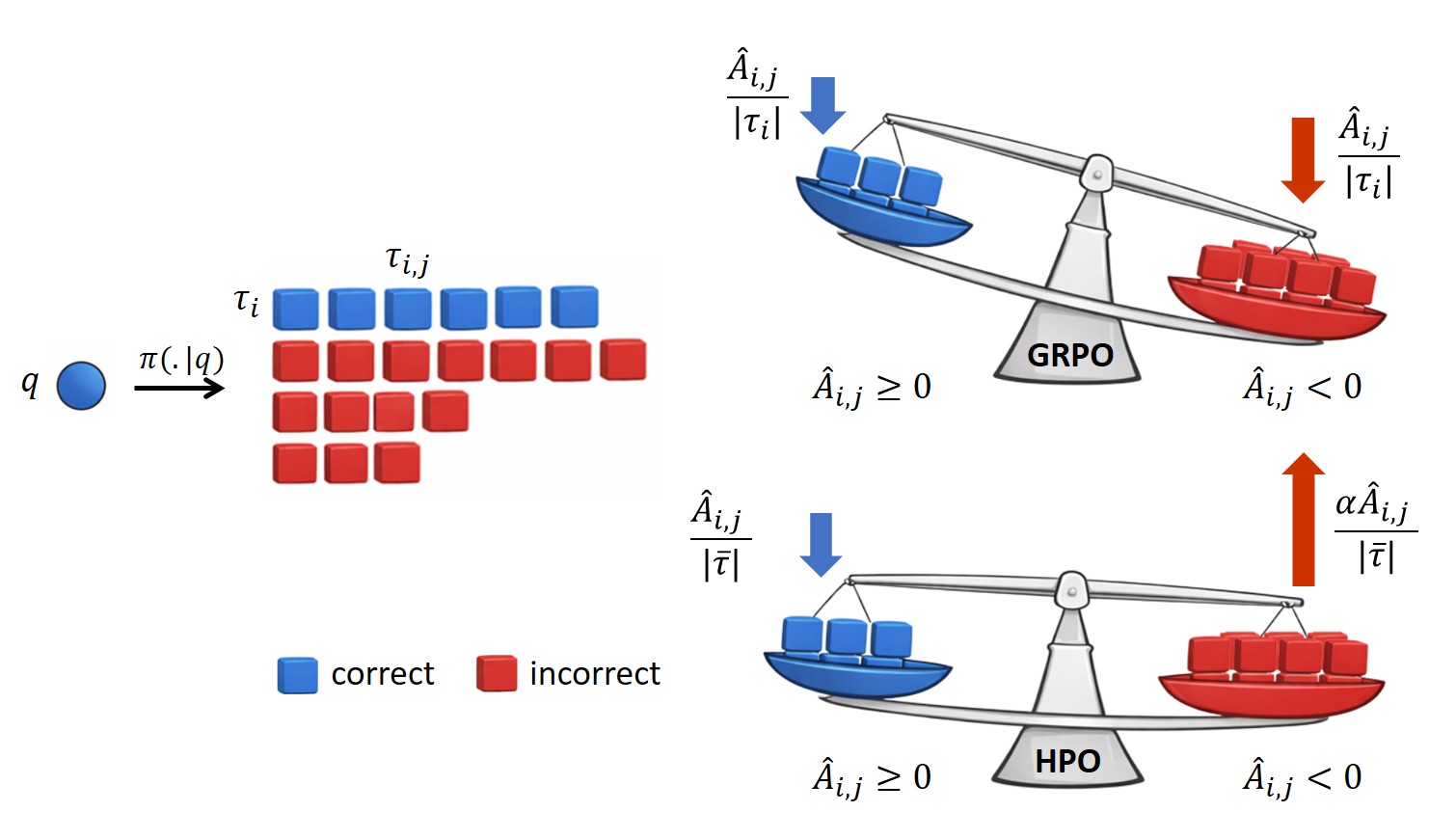}
  \caption{Conceptual view of HPO.  
    In GRPO (top), advantages are scaled by $1/|\tau_i|$ (per-response length), causing short correct responses to dominate while long incorrect ones are weakly penalized. HPO (bottom) replaces per-response length normalization with mean-length normalization and applies asymmetric hysteretic weight $\alpha$, favoring reliable positive advantages while damping uncertain negative ones, yielding more stable and compute-efficient learning.}
    \label{fig:hpo-concept}
\end{figure}

Large Language Models (LLMs) have recently demonstrated strong capabilities in reasoning, dialogue, and instruction following when fine-tuned with \ac{RL} from human or AI feedback \cite{shao2024deepseekmath, DeepSeekAIReport25, QwenReport25, ouyang2022traininglanguagemodelsfollow, rafailov2023direct, yu2025dapoopensourcellmreinforcement}. Among RL-based approaches, \ac{GRPO} has emerged as an effective framework for aligning LLMs \cite{shao2024deepseekmath}. By comparing model outputs within a group of responses to the same prompt, \ac{GRPO} provides a simple alternative to value-based policy optimization. However, GRPO remains sensitive to batch composition, rollout count, and normalization choices, making stable and efficient training challenging.

In addition, GRPO suffers from sparse-rewards \cite{feng2026rewardmap, deng2026densegrpo}. In many reasoning and diagnostic tasks, early policies solve only a small fraction of sampled prompts. As a result, successful responses are rare, and many rollouts fall below the group mean and receive negative advantages. Under limited rollout budgets, the sign distribution of advantages can become highly skewed and high-variance. In this regime, symmetric treatment of positive- and negative-advantage samples can overemphasize abundant failures, potentially suppressing noisy but partially useful reasoning trajectories. A second issue is structural. GRPO averages token-level losses by the length of each response. This response-level normalization makes the aggregate contribution of a response inversely proportional to its length: short positive-advantage responses receive larger per-token aggregate weight, while long negative-advantage responses are penalized less per response. The resulting optimization pressure is therefore not only reward-dependent but also length-dependent \cite{liu2025understanding}. This interaction is particularly relevant for language-model reasoning, where response length often reflects search, explanation, or intermediate computation.

We address these issues with \emph{Hysteretic Policy Optimization} (\HPO), a minimal modification of the \GRPO{} objective. The term hysteretic follows prior RL work that applies asymmetric learning from positive and negative update signals \cite{matignon2007hysteretic}; our setting differs in that the asymmetry is applied to response-level advantages in GRPO-style LLM fine-tuning. \HPO{} has two components: (i) asymmetric advantage weighting, where positive-advantage samples keep full weight and negative-advantage samples are down-weighted by a hysteretic factor $\alpha\in[0,1]$; and (ii) mean-length normalization, where the per-response length divisor is replaced with a batch-level mean response length. The first component targets negative-advantage dominance in sparse-reward regimes, while the second removes inverse per-response length scaling from the loss normalization. We further introduce \emph{Adaptive HPO} (\AHPO), which sets $\alpha$ from the observed fraction of positive and negative advantages in the current batch. \AHPO{} is motivated by a simple scalar balance criterion: when negative-advantage responses are more frequent than positive-advantage responses, their individual weight should be reduced so that positive and negative update contributions remain comparable. This sign-frequency rule is a lightweight heuristic rather than a direct gradient-norm estimator, and it avoids tuning a fixed $\alpha$ when batch-level sign statistics are sufficiently reliable.

Empirically, we find that \HPO{} and \AHPO{} improve reward per update in sparse-reward regimes while preserving the simplicity of \GRPO{}. \AHPO{} removes the need for manual tuning of the hysteretic weight and tracks the intermediate regime between positive-only and fully symmetric updates. Across TeleLogs and Countdown, this yields smoother training and stronger early learning under matched rollout budgets, positioning \HPO{} as a practical, low-overhead modification for compute-efficient LLM fine-tuning with sparse verifiable rewards. Our contributions are:
\begin{itemize}
    \item  we identify the dominance of negative advantages and response-level normalization as two interdependent failure modes of GRPO-style learning in the presence of sparse rewards;
    \item we propose HPO, a one-parameter interpolation between positive-only and symmetric policy-gradient updates with mean-length normalization;
    \item we introduce A-HPO, a lightweight adaptive rule to dynamically adjust the hysteretic weight based on batch-level sign statistics;
    \item we conduct ablation studies showing that the best behaviour lies between the positive-only and fully symmetric extremes.
\end{itemize}

\section{Background and Motivation}

\subsection{Sparse outcome rewards.}
Let $q$ denote a prompt and $a$ its ground-truth answer sampled from a training dataset with distribution $p_{\mathcal{D}}$. A policy $\pi_\theta$ samples a trajectory $\tau\sim\pi_\theta(\cdot\mid q)$, from which a parser\footnote{In general, $\rho(\cdot)$ is a parsing function extracting the final answer enclosed within a \textbackslash boxed\{\} tag that the LLM is prompted to provide.} $\rho(\tau)$ extracts the final answer. In the sparse-reward setting considered here, the reward is binary:
\begin{equation}
    R(\tau,a)=\mathbbm{1}\{\rho(\tau)=a\}.
\end{equation}
The RL objective is to maximize expected reward:
\begin{equation}\label{eq:objective-function}
    {\cal J}(\theta)=\E_{(q,a)}\E_{\tau\sim\pi_\theta(\cdot\mid q)}[R(\tau,a)].
\end{equation}

\subsection{Group Relative Policy Optimization}

In GRPO, the objective function Eq. \eqref{eq:objective-function} is replaced with 
\begin{align}\label{eq:GRPO}
    \mathcal{J}_{\rm GRPO}(\theta) = &\mathbb{E}_{\mathcal{X}}\left[\frac{1}{N} \sum_{i=1}^N \frac{1}{|\tau_i|} \sum_{j=1}^{|\tau_i|} \ell_{i,j}(\theta)\right],
\end{align}
where $\mathcal{X}=\{(q,a)\sim p_\cal{D}, \{\tau_i\}_{i=1}^N\sim \pi_{\rm old}(\cdot|q)\}$,
\begin{align}
\ell_{i,j}(\theta) &= \min\big(\eta_{i,j}(\theta)\hat{A}_{i,j}, \nonumber \\ &\clip(\eta_{i,j}(\theta), 1-\epsilon, 1+\epsilon)\hat{A}_{i,j} \big)
\end{align}
is the \ac{PPO} loss \cite{schulman2017proximal}, and
\begin{align}
    \eta_{i,j}(\theta) = \frac{\pi_\theta(\tau_{i,j} | q, \tau_{i,< j})}{\pi_{\rm old}(\tau_{i,j} | q, \tau_{i,< j})}
\end{align}
is the probability ratio between the old and current policy, and $\epsilon\in[0,1]$ is a clipping coefficient preventing large policy updates, which may cause instabilities.

In GRPO, the advantage is estimated on the basis of the group response--the rewards of $N$ trajectories sampled for each question. Let $R_i = R(\tau_i, a_i)$ denote the scalar reward of response $\tau_i$.
GRPO computes a group-normalized response-level advantage
\begin{equation}
    \label{eq:adv}
    \hat A_i = \frac{R_i-\mathrm{mean}(\mathbf R)}{\mathrm{std}(\mathbf R)},
\end{equation}
where $\mathbf R=\{R_1,\ldots,R_N\}$ denotes the reward vector of the group. 
This advantage is then broadcast to all response tokens, i.e., $\hat A_{i,j}=\hat A_i$ \cite{liu2025understanding}.

This formulation compares responses within each question group rather than to a fixed baseline, reducing the dependence on a reference model serving as a value model.

\paragraph{Two coupled failure modes.} \ac{GRPO} suffers from two biases that affect learning performance: 
\begin{enumerate}
    \item response-level length bias, induced by normalizing advantages by $|\tau_{i}|$. As a result, when advantages are positive, shorter answers receive proportionally stronger positive gradients, biasing the model toward brevity. When advantages are negative, longer responses contribute smaller negative gradients, reducing the penalty for verbose incorrect answers; and
    \item question-level difficulty bias, arising from per-group standardization. Both are exacerbated under small batch sizes and few rollouts.
\end{enumerate}

\paragraph{Observation: sparse rewards induce sign imbalance.}
For a prompt group, define
\[
p_+ = \Pr(\hat A_{i} > 0), \qquad
p_- = \Pr(\hat A_{i} < 0).
\]
In sparse-reward settings, successful responses are rare, so only a small fraction of sampled responses lies above the group mean. Consequently, $p_- > p_+$, and the GRPO update contains many more negative-advantage terms than positive-advantage terms. This sign imbalance is especially pronounced when the initial policy has low pass rate and the number of rollouts per prompt is small. Appendix \ref{app:sign-imbalance} provides more details on the sign imbalance observed in sparse-reward regime. 

This observation motivates hysteretic weighting: rather than treating the numerous negative-advantage samples symmetrically with the rare positive samples, HPO down-weights negative-advantage updates while preserving positive-advantage reinforcement.

\section{Hysteretic Policy Optimization}\label{sec:HPO}

\subsection{Fixed HPO}

We introduce \ac{HPO}, a modification of \ac{GRPO} that employs hysteretic learning to achieve stable updates in the sparse-rewards regime. \ac{HPO} incorporates the following key ideas:
\begin{itemize}
    \item First, to prevent question-level difficulty bias, following \cite{liu2025understanding}, we remove the standard deviation Eq. \eqref{eq:adv} and instead define:
    \begin{equation}\label{eq:adv-estimate}
        \hat{A}_{i} =  {R_i-\mathrm{mean}(\mathbf R)}
    \end{equation}
    \item Second, we replace per-response length normalization by mean-length normalization. Let
    \begin{equation}
        |\bar\tau|=\mathbb{E}_{\cal X}\!\left[\frac{1}{N}\sum_{i=1}^{N}|\tau_i|\right]
    \end{equation}
    be the mean response length. \HPO{} uses $|\bar\tau|$ as the normalization constant for all responses in the batch, so the gradient scale remains approximately stable without making the update magnitude inversely proportional to each individual response length.
    \item Third, we scale policy gradient updates asymmetrically, depending on the sign of the advantage $\hat{A}_{i}$. More specifically, we define a scale factor $w_{i,j}$:
    \begin{equation}\label{eq:hysteretic-learning}
        w_{i}=\left\{ 
      \begin{array}{ c l }
        1& \quad \textrm{if }  \hat{A}_{i} \geq 0, \\
        \alpha                 & \quad \textrm{otherwise.}
      \end{array}
    \right.
    \end{equation}
    where $\alpha\in[0,1]$.

\end{itemize}
The resulting objective function is:
\begin{align}
    \mathcal{J}_{\rm HPO}(\theta) = &\mathbb{E}_{\cal{X}}\left[\frac{1}{N|\bar{\tau}|} \sum_{i=1}^N \sum_{j=1}^{|\tau_i|} w_{i}\ell_{i,j}(\theta)\right].
    \label{eq:hpo-objective}
\end{align}
HPO is not a new complex optimizer; it is a one-parameter continuum between two known extremes. The experiments show that both extremes are suboptimal in sparse-reward LLM RL: $\alpha=0$ collapses to short positive-only guessing, while $\alpha=1$ can underperform by treating abundant negative samples symmetrically. The useful regime lies in between, and A-HPO tracks it automatically.

\subsection{Adaptive HPO}
\label{sec:adaptive-hpo}

Fixed HPO uses a constant negative-update weight $\alpha\in[0,1]$. While simple, the appropriate amount of hysteresis depends on the training regime. Early in sparse-reward training, positive-advantage samples are rare, and the update can be dominated by negative-advantage terms. Later, as the policy improves, the advantage-sign distribution becomes more balanced and a more symmetric update is desirable. This motivates an adaptive choice of
$\alpha$ based on the observed sign statistics of the current batch.

Let $\mathcal I_+ = \{i: \hat A_{i} > 0\}$ and ${\mathcal I_- = \{i: \hat A_{i}< 0\}}$ denote the positive- and negative-advantage index sets. We define the corresponding expected gradient components as
\begin{align}
G_+
&=
\mathbb{E}_{\mathcal X}\!\left[
\frac{1}{N|\bar{\tau}|}
\sum_{i\in\mathcal I_+}
\sum_{j=1}^{|\tau_i|}
\nabla_\theta \ell_{i,j}(\theta)
\right],
\nonumber\\
G_-
&=
\mathbb{E}_{\mathcal X}\!\left[
\frac{1}{N|\bar{\tau}|}
\sum_{i\in\mathcal I_-}
\sum_{j=1}^{|\tau_i|}
\nabla_\theta \ell_{i,j}(\theta)
\right].
\label{eq:positive-negative-gradient-components}
\end{align}
With hysteretic weighting, the expected HPO update can be written
schematically as
\begin{equation}
\nabla_\theta \mathcal{J}_{\rm HPO}(\theta)
=
G_+ + \alpha G_- .
\label{eq:hpo-gradient-decomposition}
\end{equation}
Here, $G_+$ and $G_-$ aggregate the positive- and negative-advantage
contributions before applying the hysteretic weight. Since $G_-$ can have a
larger magnitude in sparse-reward regimes, a natural balance criterion is to
choose $\alpha$ such that the positive and negative contributions have
comparable expected magnitudes:
\begin{equation}
\|G_+\| \approx \|\alpha G_-\|.
\label{eq:magnitude-balance}
\end{equation}
This gives the magnitude-balanced negative-update weight
\begin{equation}
\alpha
\approx
\frac{\|G_+\|}{\|G_-\|}.
\label{eq:balanced-alpha-general}
\end{equation}

Directly estimating the gradient norms in
Equation~\eqref{eq:balanced-alpha-general} is costly and noisy during
training. We therefore approximate the ratio using the sign frequencies of the
current batch. Let
\begin{equation}
\hat p_+
=
\frac{|\mathcal I_+|}{|\mathcal I_+|+|\mathcal I_-|},
\qquad
\hat p_-
=
1 - \hat p_+.
\label{eq:sign-frequency-estimates}
\end{equation}
If the average gradient magnitudes are of comparable scale across positive-
and negative-advantage samples, then the relative magnitude of $G_+$ and
$G_-$ is dominated by the number of terms in each set:
\begin{equation}
\frac{\|G_+\|}{\|G_-\|}
\approx
\frac{
\hat p_+\,\mathbb{E}[\|\nabla_\theta \ell_{i,j}\|\mid \hat A_{i}>0]
}{
\hat p_-\,\mathbb{E}[\|\nabla_\theta \ell_{i,j}\|\mid \hat A_{i}<0]
}
\approx
\frac{\hat p_+}{\hat p_-}.
\label{eq:sign-frequency-ratio}
\end{equation}

\AHPO{} implements this approximation with clipping:
\begin{equation}
\alpha_{\rm adaptive}
=
\operatorname{clip}
\left(
\frac{\hat p_+}{\hat p_-+\epsilon},
\alpha_{\min},
1
\right),
\label{eq:adaptive-alpha}
\end{equation}
where $\epsilon>0$ prevents division by zero and $\alpha_{\min}$ avoids discarding negative samples entirely. The A-HPO objective is identical to Eq.~\eqref{eq:hpo-objective}, except that the fixed $\alpha$ is replaced by $\alpha_{\rm adaptive}$ at each update.

This adaptive rule has the desired limiting behavior. In the sparse-reward regime, $\hat p_+\ll \hat p_-$, A-HPO strongly down-weights negative-advantage updates and reduces the influence of abundant failures on the update. As training progresses and the policy produces successful responses more frequently, $\hat p_+/\hat p_-$ increases toward one, causing \AHPO{} to recover a nearly symmetric HPO update. Thus, A-HPO applies hysteresis primarily
when the reward signal is sparse and automatically relaxes it when the batch statistics become less imbalanced.

\paragraph{Remark.}
We do not claim that $\hat p_+/\hat p_-$ is an unbiased estimator of gradient-norm ratio. Rather, it is a lightweight control signal for the dominant sparse-reward regime. Section~\ref{sec:grad_imbalance_proxy} validates post hoc that the resulting updates are better balanced under a clipped-surrogate contribution diagnostic. Additional details on A-HPO adaptive rule are provided in Appendix~\ref{app:balance-criterion}.

\section{Related Work}
\label{sec:related-work}

\paragraph{Group-based policy optimization.}
Recent work has revisited RL objectives for LLMs through group-relative advantages, importance weighting, and sequence structure. 
\ac{GSPO} replaces token-level importance ratios with sequence-level ratios, performing clipping and optimization at the sequence level to improve stability, especially for long responses and \ac{MoE} models \cite{zheng2025group}. 
\ac{SAPO} instead uses temperature-controlled soft gates on token-level importance ratios, with asymmetric temperatures for positive and negative updates to account for their different roles in LLM RL \cite{gao2025soft}. 
DrGRPO addresses response-level length bias and question-level difficulty by modifying the GRPO normalization, improving token efficiency and reducing long incorrect generations \cite{liu2025understanding}. 
Our work is complementary: \HPO{} keeps the GRPO-style group-relative structure but introduces explicit hysteretic weighting of negative-advantage responses together with mean-length normalization.

\paragraph{Rejection-based and selective-update alignment.}
Another line of work studies whether negative samples should be used at all. 
\ac{RAFT} selects high-reward candidate responses via rejection sampling and trains on them with supervised fine-tuning, giving a positive-only alternative to policy-gradient RL \cite{dong2023raft}. 
RAFT++ and Reinforce-Rej further show that filtering uninformative groups, such as all-correct or all-incorrect response sets, can improve stability and \ac{KL} efficiency in reasoning tasks \cite{xiong2025minimalist}. 
These methods highlight the value of selective use of negative evidence. 
\HPO{} can be viewed as a continuous alternative to hard filtering: as $\alpha\to0$, it approaches positive-only learning (RAFT-like), while as $\alpha\to1$, it approaches a symmetric DrGRPO-like policy-gradient update. 
Our experiments show that both extremes can be suboptimal in sparse-reward regimes, and that intermediate hysteresis retains useful negative feedback without allowing abundant negative-advantage samples to dominate.

\paragraph{Sparse and dense feedback.}
A complementary line of work addresses sparse rewards by providing denser supervision signals. 
For example, \cite{cao2024enhancing} use a language-model critic to generate intermediate feedback and convert it into token- or span-level rewards for RL training. 
Similarly, self-distillation approaches such as \ac{SDPO} leverage textual feedback, including error traces, judge comments, or explanations to distill feedback-conditioned predictions into the policy \cite{hubotter2026sdpo}. Recent work also studies objective-level biases in GRPO-style RL. \cite{he2025rewardingunlikelyliftinggrpo} show that GRPO can sharpen the base-model distribution and neglect rare correct trajectories in theorem proving, motivating reward reweighting for unlikely successes. Our setting is different: we assume only scalar outcome rewards and ask how to make \ac{GRPO}-style RL more stable when dense critic feedback, process supervision, or textual feedback is unavailable.

\section{Experiments}
\label{sec:experiments}

We evaluate whether \HPO{} and \AHPO{} improve stability and sample efficiency in sparse-reward RL under matched rollout budgets. The main experiments use two tasks.

\paragraph{TeleLogs.}
TeleLogs is a root-cause-analysis task over high-dimensional 5G network contexts~\citep{sana2025reasoning}. Each instance contains a symptom and network measurements, and the model must identify the underlying fault. Compared with Countdown, TeleLogs has richer context, multiple plausible explanations, and a more ambiguous optimization landscape. We follow a two-stage setup: supervised fine-tuning (SFT) followed by RL.

\paragraph{Countdown.}
Countdown is an arithmetic construction task in which the model must combine a given set of numbers with arithmetic operations to reach a target value~\cite{tinyzero, gandhi2024streamsearchsoslearning}. We evaluate an easier 3-number setting and a harder 4-number setting.

\paragraph{Setup.}
We set the base learning rate to $\mu=10^{-6}$, and the batch size to $16$ unless otherwise specified. Here, \textbf{batch size refers to the number of prompts per optimization batch}; each prompt uses $N=8$ rollouts during training and $N=4$ during evaluation. We use PPO clip $\epsilon=0.2$, and empirically set $\alpha_{\min}=0.4$ in our evaluations. Decoding uses temperature $1$ for training and $0.6$ for testing. We set top-p to $0.95$, max tokens to $4096$ (same across methods). We also provide implementation guidelines/code in the supplementary material for VERL integration (see Appendix~\ref{app:implementation}).

\paragraph{Models.} We finetune \textbf{Qwen2.5-1.5B-Instruct} on TeleLogs. 
On the countdown task, we finetune \textbf{Qwen2.5-1.5B-Instruct}, \textbf{Llama3.1-3B-Instruct} and \textbf{Qwen2.5-7B-Instruct} on Countdown. The initial pass@1 of all models is $\ll 0.5$, placing them in the sparse-reward regime studied in this work. Additional training results are provided in Appendix~\ref{app:large-models} for Qwen2.5-7B-instruct and Qwen2.5-32B-instruct on TeleLogs. We do not include them in the main paper as they are outside the sparse-reward regime focus. All models are trained for 10 epochs using VERL~\cite{sheng2024hybridflow}.

\begin{figure}[t]
    \centering
    \includegraphics[width=0.75\textwidth]{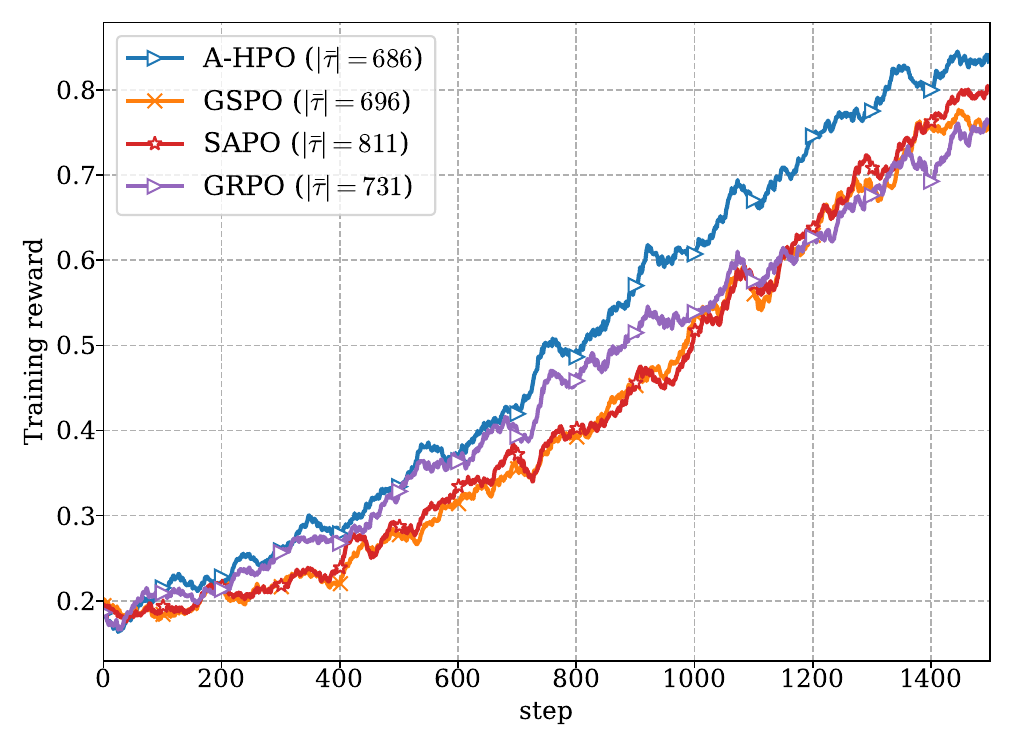}
    \caption{\AHPO{} compared with \GRPO{}, \GSPO{}, and \SAPO{} under the same training budget (model: Qwen2.5-1.5B-Instruct, dataset: TeleLogs). The legend reports each method's average response length over training. \AHPO{} reaches higher reward earlier and finishes above all baselines; at the final checkpoint it improves over \GRPO{} by a large margin and remains ahead of the stronger \GSPO{} and \SAPO{} baselines with comparable response length.}
    \label{fig:main-baselines}
\end{figure}

\subsection{Comparison with GRPO and its variants}

On TeleLogs with Qwen2.5-1.5B-Instruct, Figure~\ref{fig:main-baselines} compares \AHPO{} with \GRPO{}, \GSPO{}, and
\SAPO{} under the same training budget. \AHPO{} achieves the best reward
throughout most of training and separates from the baselines early. At the final checkpoint, \AHPO{} reaches approximately 0.84 reward, compared with 0.80 for \SAPO{}, 0.76 for \GSPO{}, and 0.73 for \GRPO{}. In relative terms, \AHPO{} improves over \GRPO{} by roughly 15\%. The gain is also visible in sample efficiency: \AHPO{} reaches around 0.60 reward near step 1000, whereas \GSPO{}, \SAPO{}, and \GRPO{} reach this level only later after an additional 200 training steps. These results show that adaptive hysteresis improves both final reward and sample efficiency, while preserving a stable response-length regime.

\begin{figure}[t]
    \centering
    \includegraphics[width=0.75\textwidth]{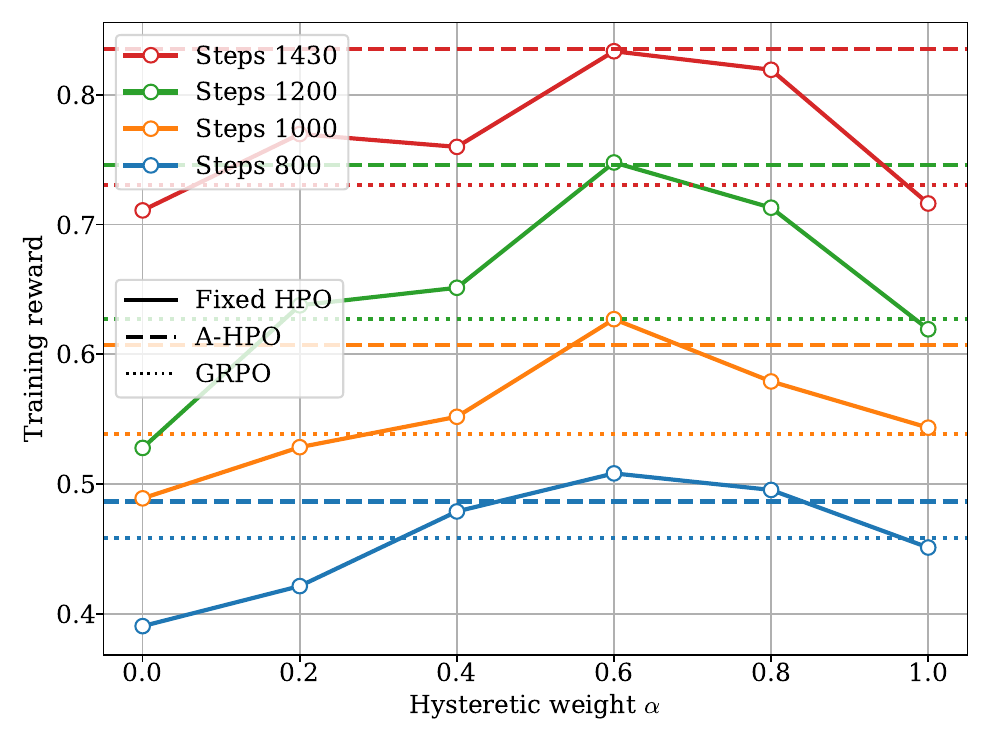}
    \caption{Sensitivity of fixed HPO to the hysteretic weight $\alpha$ (model: Qwen2.5-1.5B-Instruct, dataset: TeleLogs). Solid curves report fixed-HPO reward for different training checkpoints. Dashed horizontal lines show A-HPO and dotted horizontal lines show GRPO at the same checkpoints. Performance is non-monotonic in $\alpha$: the positive-only limit $\alpha=0$ discards useful negative feedback, while the symmetric case $\alpha=1$ underperforms in the negative-dominated regime.. Intermediate values, especially $\alpha\approx0.6$, perform best. A-HPO closely matches the best fixed-HPO region across checkpoints while avoiding manual tuning of $\alpha$.}
    \label{fig:alpha-sensitivity}
\end{figure}

\subsection{Sensitivity to the hysteretic weight}
On TeleLogs with Qwen2.5-1.5B-Instruct, Figure~\ref{fig:alpha-sensitivity} and Table~\ref{tab:alpha-length-final} study the effect of the fixed hysteretic weight $\alpha$ and compare it with \AHPO{} and \GRPO{}. The results show a clear non-monotonic dependence on $\alpha$. At one extreme, the positive-only setting $\alpha=0$ (RAFT-like) is consistently suboptimal: at the final checkpoint it obtains 0.71 reward and collapses to a mean length of only 16 tokens. This indicates that negative-advantage samples should not be discarded entirely, since they still provide useful contrastive signal. At the other extreme, the symmetric setting $\alpha=1$ (DrGRPO-like), also underperforms, reaching only 0.72 reward despite maintaining a long mean length of 696 tokens. Thus, the gains are not explained solely by the mean-length normalization; asymmetric damping of negative advantages is necessary. The best fixed settings lie in an intermediate regime, with $\alpha=0.6$ achieving 0.83 reward and a mean length of 653 tokens. \AHPO{} matches this tuned regime without requiring an $\alpha$ sweep, reaching the best final reward of 0.84 with a mean length of 686 tokens, close to \GRPO{}'s 729 tokens but with a large accuracy improvement over \GRPO{}'s 0.73 reward. These results support the central design of \AHPO{}: it automatically applies strong hysteresis when negative updates dominate, while preserving enough negative signal to avoid the degenerate short-output behavior observed in the positive-only limit.

Additional ablations in Appendix~\ref{app:hpo-variants} study batch-size effects and diagnostic variants. In particular, N-HPO isolates adaptive hysteresis under GRPO's original length normalization. A-HPO improves over N-HPO, indicating that mean-length normalization contributes beyond hysteresis; N-HPO improves over GRPO, indicating that hysteresis itself is beneficial.

\begin{table}[!tb]
\centering
\small
\begin{tabular}{lcc}
\toprule
\textbf{Method} & \textbf{Reward} & \textbf{Mean length} \\
\midrule
\HPO, $\alpha=0.0$ & 0.71 & 16 \\
\HPO, $\alpha=0.2$ & 0.77 & 225 \\
\HPO, $\alpha=0.4$ & 0.76 & 471 \\
\HPO, $\alpha=0.6$ & 0.83 & 653 \\
\HPO, $\alpha=0.8$ & 0.82 & 715 \\
\HPO, $\alpha=1.0$ & 0.72 & 696 \\
\AHPO{} & \textbf{0.84} & 686 \\
\GRPO{} & 0.73 & 729 \\
\bottomrule\\
\end{tabular}
\caption{
Final-checkpoint evaluation reward and mean response length for HPO under different hysteretic weights $\alpha$. The positive-only limit $\alpha=0$ produces very
short outputs and underperforms, showing that HPO gains are not explained by
trivial answer shortening. The best fixed settings and A-HPO achieve the highest reward while maintaining response lengths comparable to GRPO.
}
\label{tab:alpha-length-final}
\end{table}

\subsection{Does A-HPO balance positive and negative contributions?}\label{sec:grad_imbalance_proxy}
A-HPO is designed to prevent the update from being dominated by the abundant negative-advantage samples that arise in sparse-reward training. Ideally, this would be verified by tracking the gradient-norm balance ${\|G_+\|}/{\|G_-\|}$ in Eq.~\eqref{eq:sign-frequency-ratio}. However, computing per-sample gradient norms is expensive. We therefore monitor a surrogate contribution-balance ratio based on the absolute PPO clipped surrogate:
\[
\tilde\rho
=
\frac{p_+\tilde m_+}{p_-\tilde m_-},
\qquad
\tilde m_\pm
=
\mathbb{E}\!\left[
|\ell_{i,j}(\theta)| \mid \hat A_{i}\gtrless0
\right].
\]
Here, $|\ell_{i,j}(\theta)|$ is the absolute scalar PPO contribution, which controls the magnitude of the clipped policy-gradient.

\begin{figure}[t]
    \centering
    \includegraphics[width=0.75\textwidth]{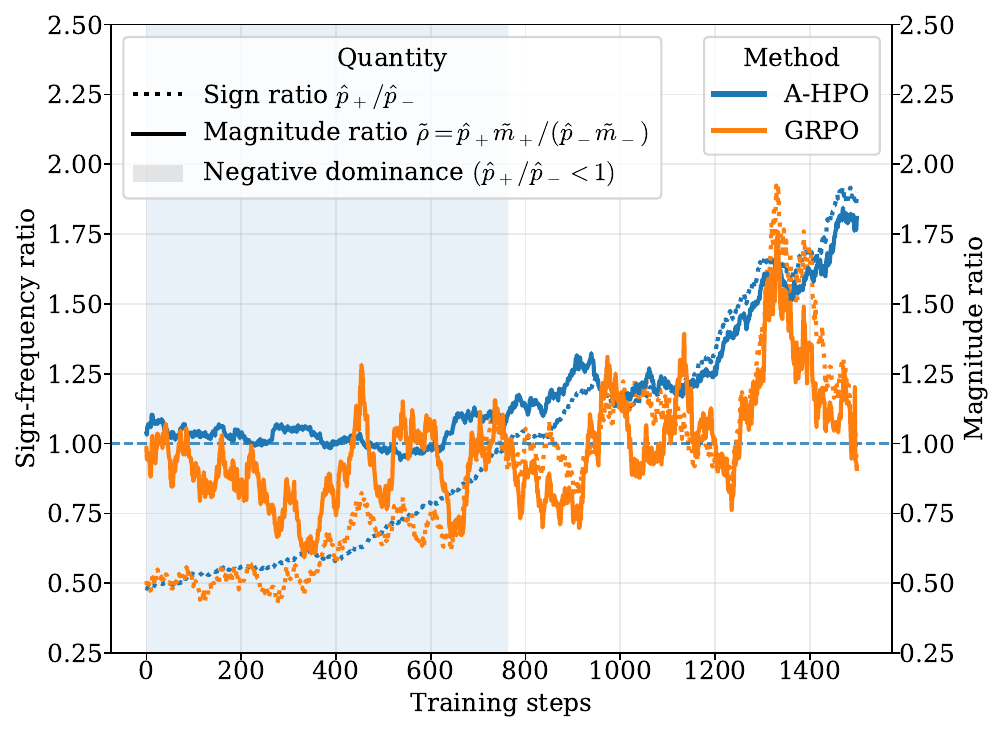}
    \caption{
    Does A-HPO balance positive and negative update contributions? Dotted curves show the sign-frequency ratio $\hat p_+/\hat p_-$ used by A-HPO. Solid curves show the surrogate contribution-balance ratio $\tilde\rho$. The shaded region marks negative dominance, $\hat p_+/\hat p_-<1$, and the horizontal dashed line marks balance. A-HPO moves out of the negative-dominated phase and maintains a surrogate balance near or above one, whereas GRPO shows a noisier and less sustained balancing behavior.
    }
    \label{fig:adaptive-ratio-diagnostics}
\end{figure}

Figure~\ref{fig:adaptive-ratio-diagnostics} shows that A-HPO indeed produces a more balanced contribution profile. Early in training, both A-HPO and GRPO operate in the negative-dominance regime, with $\hat p_+/\hat p_-<1$, confirming that sparse rewards initially generate many more negative than positive advantage terms. Under A-HPO, this imbalance is progressively corrected: the sign ratio crosses the balance threshold and the surrogate balance ratio $\tilde\rho$ remains close to, and later rises above, one. This indicates that A-HPO does not merely increase the frequency of positive samples; it also keeps their surrogate contribution comparable to the negative side. In contrast, GRPO shows a noisier contribution profile and remains closer to the balance threshold near the end of training. Thus, A-HPO acts as intended: it damps negative updates when they dominate, then relaxes the damping as the policy enters a less sparse regime, yielding a smoother transition from negative-dominated to balanced or positive-supported updates.

\subsection{Result on the Countdown reasoning across model scales}

\begin{table*}[t]
\centering
\small
\begin{tabular}{llrrrr}
\toprule
\multirow{2}{*}{\textbf{Model}} 
& \multirow{2}{*}{\textbf{Task}} 
& \multicolumn{2}{c}{\textbf{200 steps}} 
& \multicolumn{2}{c}{\textbf{1000 steps}} \\
\cmidrule(lr){3-4}\cmidrule(lr){5-6}
& & \textbf{\AHPO{}} & \textbf{\GRPO{}} & \textbf{\AHPO{}} & \textbf{\GRPO{}} \\
\midrule
\multirow{2}{*}{Qwen2.5-7B-Instruct}
& Countdown-3 & \textbf{0.94} & 0.93 & \textbf{0.95} & 0.94 \\
& Countdown-4 & \textbf{0.63} & 0.58 & \textbf{0.65} & 0.58 \\
\midrule
\multirow{2}{*}{Llama3.1-3B-Instruct}
& Countdown-3 & \textbf{0.86} & 0.78 & 0.91 & \textbf{0.93} \\
& Countdown-4 & \textbf{0.54} & 0.40 & \textbf{0.58} & \textbf{0.58} \\
\midrule
\multirow{2}{*}{Qwen2.5-1.5B-Instruct}
& Countdown-3 & \textbf{0.47} & 0.37 & \textbf{0.51} & 0.47 \\
& Countdown-4 & 0.17 & \textbf{0.19} & \textbf{0.23} & 0.20 \\
\bottomrule\\
\end{tabular}
\caption{Countdown evaluation reward across model scales and difficulty levels. Countdown-3 and Countdown-4 denote instances with three and four input numbers, respectively. \AHPO{} provides the largest gains in sparse and low-budget regimes, especially on harder Countdown-4 instances and at 200 training steps. On saturated settings, such as Qwen2.5-7B Countdown-3, both methods perform similarly.}
\label{tab:countdown-results}
\end{table*}

Table~\ref{tab:countdown-results} evaluates \AHPO{} and \GRPO{} on Countdown across three instruction-tuned backbones and two difficulty levels. The results highlight the regime in which adaptive hysteresis is most useful: early training, smaller models, and harder sparse-reward instances. On the easier Countdown-3 setting, large models are already close to saturation: for Qwen2.5-7B, \AHPO{} and \GRPO{} differ by only one point at both 200 and 1000 steps. In contrast, on the harder Countdown-4 setting, \AHPO{} gives consistent gains for the same 7B model, improving reward from 0.58 to 0.63 at 200 steps and from 0.58 to 0.65 at 1000 steps. The benefit is even clearer in the low-budget regime. For Llama3.1-3B, \AHPO{} improves 200-step reward by $+0.08$ on Countdown-3 and by $+0.14$ on Countdown-4; for Qwen2.5-1.5B, it improves Countdown-3 by $+0.10$ at 200 steps and remains better at 1000 steps ($0.51$ vs. $0.47$). These gains show that \AHPO{} accelerates learning when successful trajectories are still rare. As training progresses or when the task becomes saturated, the gap naturally narrows: \GRPO{} slightly exceeds \AHPO{} on Llama3.1-3B Countdown-3 at 1000 steps, and both methods tie on Llama3.1-3B Countdown-4. Overall, A-HPO is most useful before saturation and under harder sparse settings; when the task is easy or the model is already strong, the advantage narrows and GRPO can match or slightly exceed it.

\section{Discussion}
\label{sec:discussion}

\paragraph{Practical recipe.}
For fixed \HPO, we recommend setting $\alpha\in[0.4, 0.8]$ with $\alpha\approx0.6$ being the best in our TeleLogs analysis. \AHPO{} is preferable when batch-level sign statistics are reliable, for example with batch size at least 16 in our TeleLogs setup. When batches are extremely small, the adaptive estimate can become noisier; fixed HPO or smoothing the adaptive factor may be preferable in some settings. We include these controls in Appendix~\ref{app:telelogs-additional}.

\paragraph{Why keep negative updates at all?}
The $\alpha=0$ setting (RAFT-like) is useful as an ablation but not as the recommended method. It removes negative evidence entirely and can collapse toward short, high-variance guesses. Intermediate hysteretic weight is better aligned with the sparse-reward setting: it treats positive samples as scarce evidence while still retaining enough negative feedback to shape the policy.

\section{Conclusion}

We proposed HPO, a minimal modification of GRPO designed to finetune LLM under sparse-reward regime. HPO balances the contribution of positive- and negative-advantage updates and replaces GRPO's response-level normalization with mean-length normalization. We further introduced A-HPO, which adapts the hysteretic factor from batch-level advantage-sign statistics
Across Countdown and TeleLogs, \HPO{} and \AHPO{} improve reward per update under matched rollout budgets, compare favorably with \GRPO, \GSPO, and \SAPO{} on TeleLogs, and avoid the positive-only length-collapse failure mode. The results support a simple principle for sparse-reward RL: negative-advantage updates should be used, but not trusted symmetrically when they are overwhelmingly frequent, especially early in the training.

\bibliographystyle{plain}

\bibliography{bibliography}

\clearpage

\appendix

\section{Additional analysis}
\section{Sign imbalance under sparse binary rewards}
\label{app:sign-imbalance}

We provide a simple calculation illustrating why sparse outcome rewards induce negative-advantage dominance. Consider a prompt group with $N$ sampled responses and binary rewards $R_i\in\{0,1\}$. Let
\begin{equation}
    p = \Pr(R_i=1)
\end{equation}
denote the success probability of the current policy, and define the centered response-level advantage
\begin{equation}
    \hat A_i = R_i - \bar R,
\qquad
\bar R = \frac{1}{N}\sum_{k=1}^{N} R_k.
\end{equation}
Define
\[
p_+ = \Pr(\hat{A}_{i} > 0), \qquad
p_- = \Pr(\hat{A}_{i} < 0),
\]
where the probability is taken over the stochastic sampling of responses and rewards. We use strict signs and treat zero-advantage responses as neutral.

\begin{figure}
    \centering
    \includegraphics[width=0.75\linewidth]{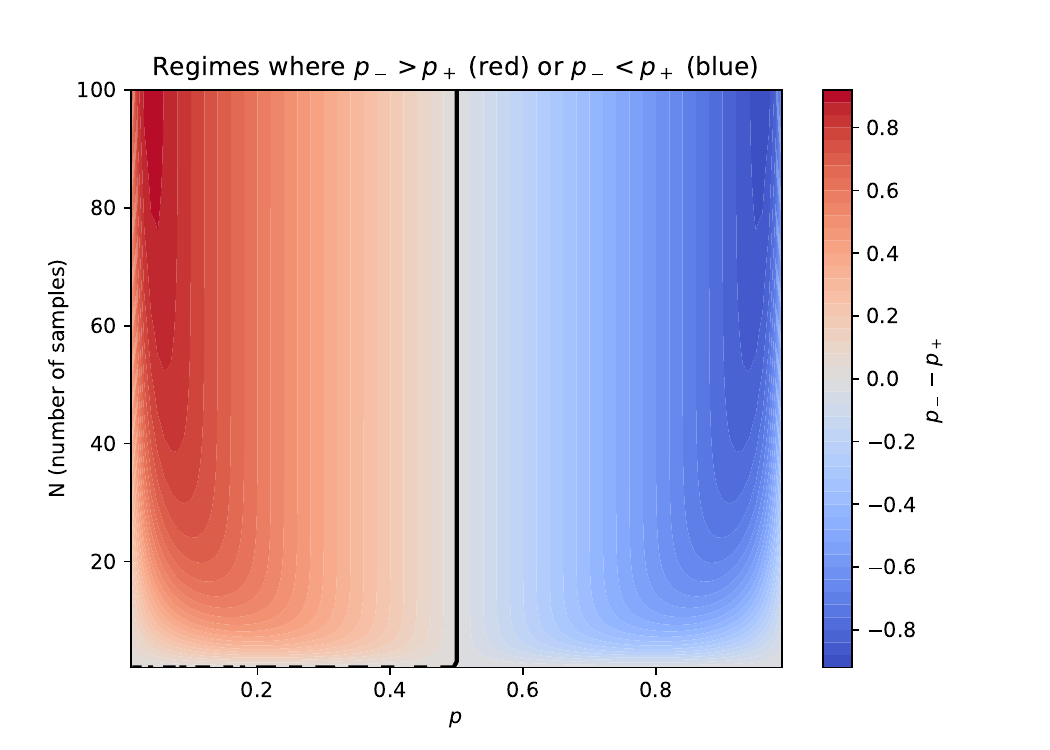}
    \caption{Negative-advantage dominance under sparse-rewards regime.}
    \label{fig:neg-advg-dominance}
\end{figure}
For binary rewards, a response has positive advantage iff it is successful and
the group is not all successful:
\[
\hat A_i>0
\quad \Longleftrightarrow \quad
R_i=1
\ \text{and}\ 
\exists k\neq i : R_k=0 .
\]
Therefore,
\begin{align}
    p_+ &= \Pr(R_i=1)\Pr(\exists k\neq i:R_k=0) \\
        &= p(1-p^{N-1}). \nonumber
\end{align}
Similarly, a response has negative advantage iff it is unsuccessful and at
least one other response in the group is successful:
\[
\hat A_i<0
\quad \Longleftrightarrow \quad
R_i=0
\ \text{and}\ 
\exists k\neq i : R_k=1 .
\]
Thus,
\begin{align}
    p_- &= \Pr(R_i=0)\Pr(\exists k\neq i:R_k=1)\\
        &= (1-p)\left(1-(1-p)^{N-1}\right). \nonumber
\end{align}
The ratio is
\begin{align}
    \frac{p_-}{p_+}
=
\frac{1-p}{p}
\cdot
\frac{1-(1-p)^{N-1}}{1-p^{N-1}}.
\end{align}
For sparse rewards, $p\ll 1/2$, this ratio is greater than one, and it grows as the success probability decreases (see Figure~\ref{fig:neg-advg-dominance}). Hence, among informative groups with mixed rewards, negative-advantage samples are more frequent than positive-advantage samples. This provides a simple explanation for the negative-dominance regime targeted by HPO and A-HPO.

\section{Magnitude-balanced adaptive factor}
\label{app:balance-criterion}

A-HPO uses the sign-frequency ratio $\hat p_+/\hat p_-$ as a lightweight adaptive estimate of the negative-update weight. Here we derive this rule from a contribution-balance criterion.

Let $G_+$ and $G_-$ denote the positive- and negative-advantage gradient components before hysteretic weighting:
\[
G_+
=
\mathbb E_{\mathcal X}
\left[
\frac{1}{N|\bar\tau|}
\sum_{i\in\mathcal I_+}
\sum_{j=1}^{|\tau_i|}
\nabla_\theta \ell_{i,j}(\theta)
\right],
\]
\[
G_-
=
\mathbb E_{\mathcal X}
\left[
\frac{1}{N|\bar\tau|}
\sum_{i\in\mathcal I_-}
\sum_{j=1}^{|\tau_i|}
\nabla_\theta \ell_{i,j}(\theta)
\right].
\]
where $\mathcal{X}=\{(q,a)\sim p_\cal{D}, \{\tau_i\}_{i=1}^N\sim \pi_{\rm old}(\cdot|q)\}$.

HPO applies weight $1$ to $G_+$ and $\alpha$ to $G_-$, giving
\[
\nabla_\theta \mathcal J_{\rm HPO}
=
G_+ + \alpha G_- .
\]
A natural balance criterion is to choose $\alpha$ so that the positive and negative components have comparable expected magnitudes:
\[
\|G_+\| \approx \|\alpha G_-\|.
\]
This yields the magnitude-balanced weight
\[
\alpha
\approx
\frac{\|G_+\|}{\|G_-\|}.
\]
Directly estimating $\|G_+\|$ and $\|G_-\|$ requires per-sample gradient norms. A-HPO therefore uses a sign-frequency approximation. If the average per-sample gradient magnitude is of comparable scale across signs, then
\[
\frac{\|G_+\|}{\|G_-\|}
\approx
\frac{
p_+\,\mathbb E[\|\nabla_\theta \ell_{i,j}\|\mid \hat A_i>0]
}{
p_-\,\mathbb E[\|\nabla_\theta \ell_{i,j}\|\mid \hat A_i<0]
}
\approx
\frac{p_+}{p_-}.
\]
This gives the adaptive rule
\[
\alpha_{\rm adaptive}
=
\operatorname{clip}
\left(
\frac{\hat p_+}{\hat p_-+\epsilon},
\alpha_{\min},
1
\right).
\]
The clipping prevents degenerate positive-only updates and recovers the symmetric update when positive and negative signs become balanced. This derivation should be interpreted as a magnitude-balancing heuristic, not as a claim of global optimality. The full policy-gradient update is vector valued and affected by clipping, optimizer state, and token correlations.

\paragraph{Zero advantages.}
When estimating $\hat p_+$ and $\hat p_-$, we ignore responses with exactly zero centered advantage:
\[
\mathcal I_+ = \{i:\hat A_i>0\},\qquad
\mathcal I_- = \{i:\hat A_i<0\}.
\]
Zero-advantage samples do not provide directional evidence for increasing or decreasing the policy probability. If all centered advantages in a batch are zero, the adaptive factor is set to $\alpha_{\rm adaptive}=1$; in this case, the policy-gradient contribution from the centered advantages is zero.

\section{Additional Algorithmic Variants}
\label{app:variants}

The main paper focuses on fixed \HPO{} and \AHPO. We keep the additional variants here to avoid obscuring the core contribution.

\subsection{Variance-aware HPO (V-HPO)}
\VHPO{} sets the hysteretic factor using group-level reward variance:
\begin{equation}\label{eq:variance-aware-learning}
    \alpha=\alpha_0\left(1-\exp\left(-\frac{\alpha_1}{\delta^2+\varepsilon}\right)\right),
\end{equation}
where $\delta = 0.5-\mathrm{std}(\mathbf R))$ denotes the distance from the group reward std to the maximum std for bounded binary rewards, which is $0.5$. This variant interprets reward variance as a confidence signal. 
In Eq. \ref{eq:variance-aware-learning}, the exponential term modulates the hysteretic weight $\alpha$ based on the group reward variance. A low variance indicates high confidence in the group-level reward signal:
the sampled responses are mostly consistent, so V-HPO uses a low hysteretic weight and strongly damps negative-advantage updates. In contrast, a higher variance indicates a more mixed or uncertain group, so V-HPO uses a larger hysteretic weight, allowing negative-advantage samples to contribute more and moving the update closer to the symmetric GRPO regime.

\subsection{Normalization effect isolation.}
N-HPO keeps GRPO's original response-level length normalization while applying the same adaptive hysteretic weights as A-HPO. This control isolates the contribution of mean-length normalization: differences between A-HPO and N-HPO reflect the effect of replacing per-response normalization by mean-length normalization, while differences between N-HPO and GRPO reflect the effect of adaptive hysteretic weighting under the original GRPO normalization.

\begin{figure}[t]
    \centering
    \begin{subfigure}[t]{0.48\textwidth}
        \centering
        \includegraphics[width=\linewidth]{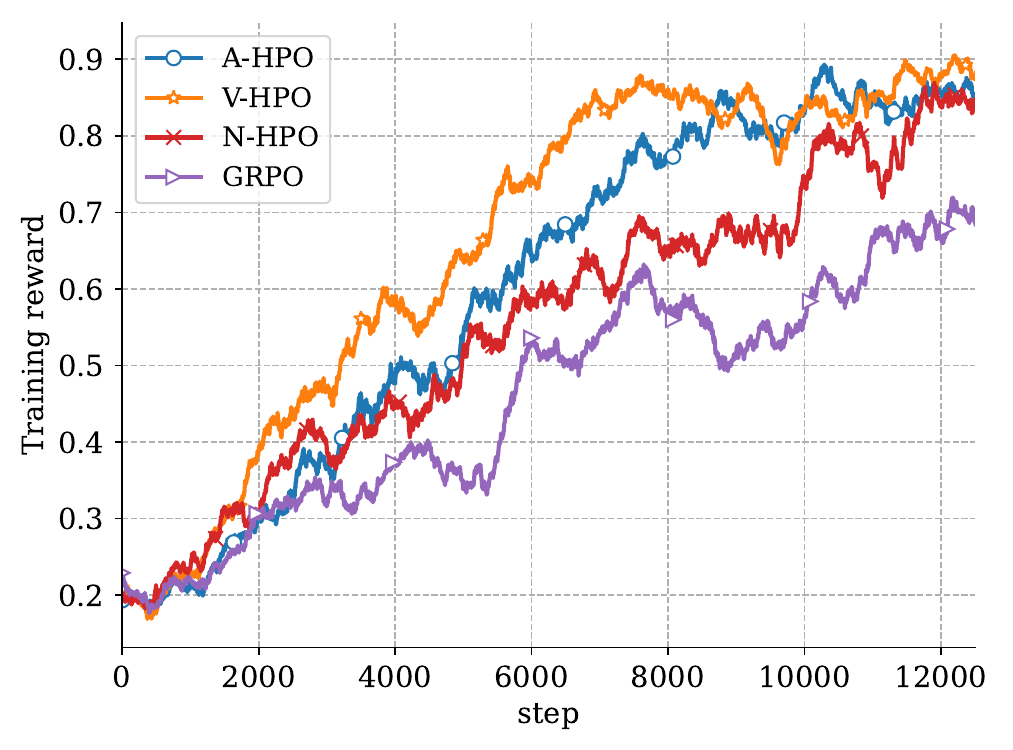}
        \caption{Batch size 1.}
    \end{subfigure}
    \hfill
    \begin{subfigure}[t]{0.48\textwidth}
        \centering
        \includegraphics[width=\linewidth]{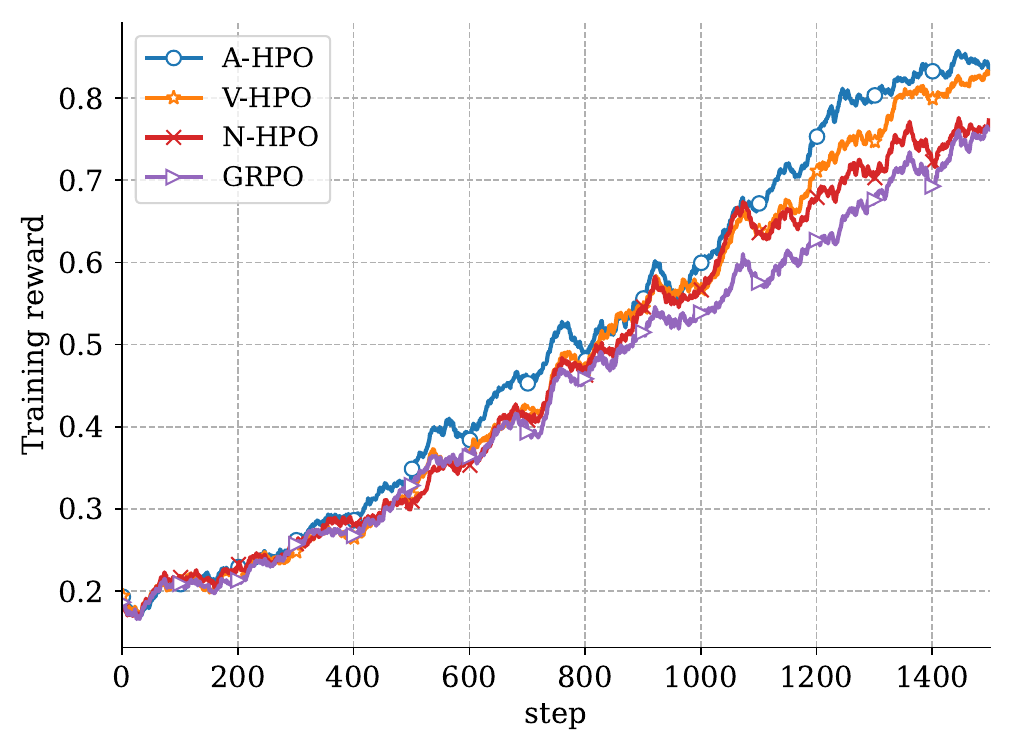}
        \caption{Batch size 16.}
    \end{subfigure}
    \caption{Ablation of HPO variants on TeleLogs under batch sizes 1 and 16 (model: Qwen2.5-1.5B-Instruct). N-HPO keeps \GRPO{}'s response-level length normalization while applying adaptive hysteretic weights, isolating the effect of mean-length normalization. V-HPO uses a variance-aware adaptive factor. Across both batch sizes, the HPO variants outperform \GRPO{}, while \AHPO{} provides the best overall stability-performance trade-off.}
    \label{fig:hpo-variants-bs}
\end{figure}

\section{Additional results on TeleLogs} \label{app:hpo-variants}

\subsection{Batch-size ablation and diagnostic variants}
\label{app:telelogs-additional}

Figure~\ref{fig:hpo-variants-bs} compares \AHPO{} with two diagnostic variants on TeleLogs under batch sizes $1$ and $16$.
It should be noted that \textbf{batch size refers to the number of prompts per optimization batch}; each prompt uses $N=8$ rollouts during training. Thus, even with batch size 1, $\alpha$ is estimated from the $N = 8$ rollouts for the prompt, not from a single response.

N-HPO keeps \GRPO{}'s response-level length normalization while applying the same adaptive hysteretic weighting as
\AHPO{}, thereby isolating the contribution of mean-length normalization. V-HPO uses a variance-aware adaptive factor. In the noisy batch size=1 regime, all HPO variants substantially outperform \GRPO{}, showing that hysteretic weighting is
especially useful when reward and sign statistics are unstable. N-HPO improves strongly over \GRPO{}, confirming that adaptive hysteresis alone is beneficial, while \AHPO{} gives a smoother and competitive final trajectory, showing the
additional stabilizing effect of mean-length normalization. With batch size=16, all methods train more smoothly, but \AHPO{} remains the strongest overall, finishing above both N-HPO and \GRPO{}. These results support the design of
\AHPO{} as the main method: adaptive hysteresis provides the core sparse-reward gain, and mean-length normalization improves stability and final performance.

\subsection{Larger TeleLogs Models}
\label{app:large-models}

Table~\ref{tab:telelogs-large-models} reports TeleLogs pass@1 for larger models of the Qwen family trained with HPO.
These runs start from SFT checkpoints whose initial pass@1 is close to 0.5 (particularly for 32B), and thus do not reflect the sparse-reward regime studied in the main paper.
We include them to demonstrate that HPO training remains stable and can further improve performance at larger model scales.

\begin{table}[h]
\centering
\small
\setlength{\tabcolsep}{6pt}
\begin{tabular}{lccc}
\toprule
\textbf{Training stage} & \textbf{1.5B} & \textbf{7B} & \textbf{32B} \\
\midrule
Base & 11.25 & 12.05 & 18.85 \\
SFT & 19.62 & 39.20 & 49.45 \\
SFT + RL (\HPO) & 80.41 & 87.00 & 95.86 \\
\bottomrule\\
\end{tabular}

\caption{TeleLogs pass@1 (\%) for larger Qwen-family models. These runs start from strong SFT checkpoints, especially the 32B model, and are therefore feasibility evidence rather than the primary sparse-regime evaluation.}
\label{tab:telelogs-large-models}
\end{table}

\begin{table}[htb]
\centering
\small
\begin{tabular}{ll}
\toprule
Hyperparameter & Value \\
\midrule
Learning rate & $10^{-6}$ \\
PPO clip & $0.2$ \\
Rollouts per prompt & $N=8$ \\
Evaluation rollouts & $N=4$ \\
Batch size & 16 prompts \\
Training temperature & 1.0 \\
Evaluation temperature & 0.6 \\
Top-p & 0.95 \\
Max response length & 4096 \\
$\alpha_{\min}$ & 0.4 \\
V-HPO ($\alpha_{0}$) & 0.4 \\
V-HPO ($\alpha_{1}$) & $10^{-2}$ \\
GSPO (clip low) & $3\times 10^{-3}$ \\
GSPO (clip high) & $3\times 10^{-4}$ \\
SAPO (positive temperature) & 1.0 \\
SAPO (negative temperature) & 1.05 \\
\bottomrule\\
\end{tabular}
\caption{Main hyperparameters used in our experiments.}
\label{tab:hyperparameters}
\end{table}

\section{Implementation details}
\label{app:implementation}

\subsection{Computational overhead}
\label{app:overhead}

HPO and A-HPO add negligible computational overhead relative to GRPO. Fixed HPO only multiplies response-level advantages by a sign-dependent scalar. A-HPO adds a batch-level count of positive and negative response-level advantages and one scalar clipping operation. No additional model forward pass, value model,
reward model, or per-sample gradient computation is required. The method is therefore compatible with existing GRPO/PPO training pipelines.

\subsection{VERL implementation}

In VERL, \HPO{} can be implemented by modifying the advantage estimator. The implementation computes centered response-level advantages, applies sign-dependent hysteretic scaling, and broadcasts the resulting advantage to tokens with the response mask. Mean-length normalization is applied in the policy loss rather than by changing the response reward.

\begin{figure*}
\includegraphics[width=1.2\textwidth]{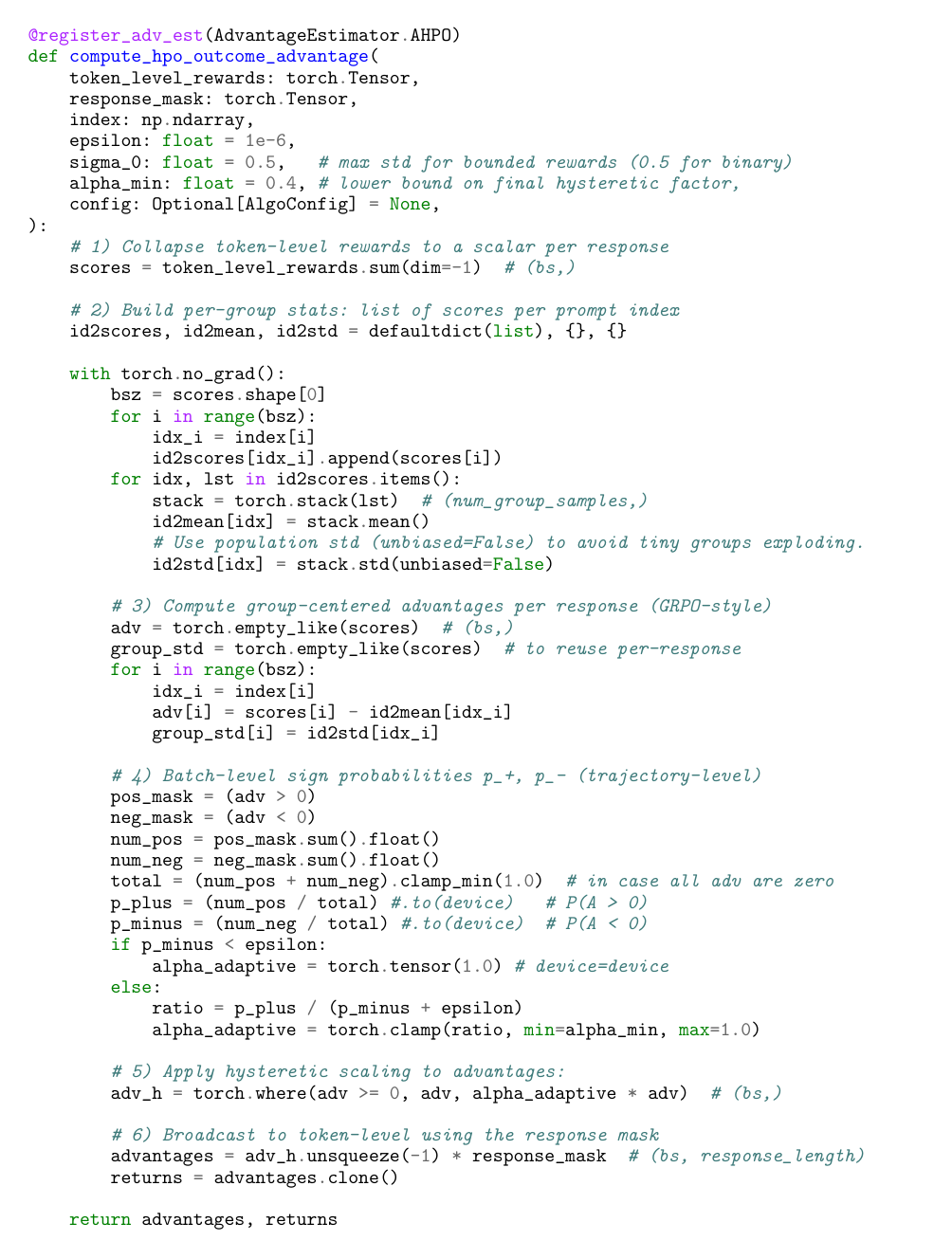}
\caption{Implementation of A-HPO's algorithm in VERL}\label{eq:adv_estimator}
\end{figure*}

In practice, the HPO algorithm can be easily integrated into VERL~\cite{sheng2024hybridflow} by registering a new advantage estimator as shown in Figure \ref{eq:adv_estimator}. Table~\ref{tab:hyperparameters} summarizes the hyperparameters used for training.

\begin{acronym}[AAAAAAAAA]
\acro{KL}{Kullback–Leibler}
\acro{PPO}{Proximal Policy Optimization}
\acro{HPO}{Hysteretic Policy Optimization}
\acro{GRPO}{Group Relative Policy Optimization}
\acro{RLHF/RLAIF}{Reinforcement Learning from Human or AI Feedback}
\acro{SAPO}{Soft Adaptive Policy Optimization}
\acro{GSPO}{Group Sequence Policy Optimization}
\acro{RAFT}{Reward rAnked FineTuning}
\acro{SDPO}{Self-Distillation Policy Optimization}
\acro{RL}{Reinforcement Learning}
\acro{MoE}{Mixture-of-Experts}
\end{acronym}

\end{document}